\documentclass[journal]{IEEEtran}
\ifCLASSINFOpdf
\else
\fi
\usepackage{times}
\usepackage{epsfig}
\usepackage{graphicx}
\usepackage{amsmath}
\usepackage{amssymb}
\usepackage{url}            
\usepackage{booktabs}       
\usepackage{amsfonts}       
\usepackage{nicefrac}  
\usepackage{enumitem}
\usepackage{lipsum}
\usepackage{microtype}      
\usepackage{graphicx}
\usepackage{multicol,multirow}
\usepackage{wrapfig}
\usepackage{color}
\usepackage{float}
\usepackage{bm}
\usepackage{dashrule}
\usepackage[table]{xcolor}
\usepackage{algorithm2e}
\usepackage{caption,mathtools}
\usepackage{subcaption}
\usepackage{listings}

\DeclareMathOperator*{\argmin}{arg\,min}

\def\x{{\mathbf x}}

\def\etal{\emph{et. al.}}

\def\ie{\emph{i.e}.}

\newcommand{\norm}[1]{\left\lVert#1\right\rVert}

\usepackage[pagebackref=true,breaklinks=true,letterpaper=true,colorlinks,bookmarks=false]{hyperref}

\usepackage[margins]{trackchanges}
\addeditor{Jay}

\begin{document}
%



\title{
Revisiting Deep Subspace Alignment for Unsupervised Domain Adaptation
}

%
%
%

\author{Kowshik Thopalli,
        Jayaraman J Thiagarajan,
        Rushil Anirudh,~and~Pavan K Turaga~\IEEEmembership{Senior Member,~IEEE} 
\thanks{Kowshik Thopalli and Pavan Turaga are with the Geometric Media Lab, Department
of Electrical and Computer Engineering, Arizona State University,
AZ, 85287, USA e-mail:kthopall@asu.edu,pturaga@asu.edu}
\thanks{Jayaraman J. Thiagarajan and Rushil Anirudh are with Lawrence Livermore National Laboratory, CA, 94550,USA.email: jjayaram@llnl.gov, anirudh1@llnl.gov}
\thanks{Manuscript received December 16, 2021;}}
%
%

\markboth{Thopalli \MakeLowercase{\textit{et al.}}: Revisiting Deep Subspace Alignment for Unsupervised Domain Adaptation}%
{Thopalli \MakeLowercase{\textit{et al.}}: Revisiting Deep Subspace Alignment for Unsupervised Domain Adaptation}
%



\maketitle

\begin{abstract}

Unsupervised domain adaptation (UDA) aims to transfer and adapt knowledge from a labeled source domain to an unlabeled target domain. 
Traditionally, subspace-based methods form an important class of solutions to this problem. Despite their mathematical elegance and tractability, these methods are often found to be ineffective at producing domain-invariant features with complex, real-world datasets. 
Motivated by the recent advances in representation learning with deep networks, this paper revisits the use of subspace alignment for UDA and proposes  a novel adaptation algorithm that consistently leads to improved generalization. In contrast to existing adversarial training-based DA methods, our approach isolates feature learning and distribution alignment steps, and utilizes a primary-auxiliary optimization strategy to effectively balance the objectives of domain invariance and model fidelity. While providing significant reduction in target data and computational requirements, our subspace-based DA performs competitively and sometimes even outperforms state-of-the-art approaches on several standard UDA benchmarks. Furthermore, subspace alignment leads to intrinsically well-regularized models that demonstrate strong generalization even in the challenging partial DA setting. Finally, the design of our UDA framework inherently supports progressive adaptation to new target domains at test-time, without requiring retraining of the model from scratch. In summary, powered by powerful feature learners and an effective optimization strategy, we establish subspace-based DA as a highly effective approach for visual recognition.
\end{abstract}


\IEEEpeerreviewmaketitle

\section{Introduction}

In the past decade, advances in computing hardware coupled with access to large amounts of labeled data have led to remarkable success of supervised deep learning in computer vision, natural language processing, and audio processing. A common assumption often made by such supervisory solutions is that training and testing data are independent and identically distributed (i.i.d.). However, in practice, this assumption seldom holds true and in such cases, we often witness a significant drop in performance~\cite{wang2018deep}. As a result, effectively generalizing to testing scenarios, characterized by unknown distribution shifts, remains a long standing challenge. This fundamental challenge has been formalized using a variety of formulations, which can be broadly categorized based on the availability of data from the \textit{source} domain, access to the parameters of the model trained on source data, and access to either labeled or unlabeled data samples from the \textit{target} domain of interest. In this paper, we focus on unsupervised domain adaptation (UDA), where one can access only unlabeled data from the target domain along with labeled data from the source domain. 

At the core of state-of-the-art domain adaptation approaches are three components that need to be synergistically trained: (i) a feature-learner implemented using deep neural networks; (ii) an appropriate alignment procedure between source and target feature distributions to achieve domain-invariance; and (iii) an optimization strategy that trades-off between alignment quality and model fidelity on the labeled source data. In this context, subspace-based domain adaptation forms a popular class of approaches~\cite{Subspace_alignment, Gopalan2011DomainAF, Gong2012GeodesicFK}, wherein the source and target domains are modeled as low-dimensional subspaces, and the alignment can be expressed as an affine transformation between the subspaces. Despite its tractability, simple subspace models are often found to be insufficient for the complex data distributions we deal with in practice~\cite{deepcoral} and hence, more sophisticated approaches have emerged. Examples include adversarial learning~\cite{cite:CVPR17ADDA,cite:CVPR18GTA}, optimal-transport~\cite{bhushan2018deepjdot} and distribution-matching~\cite{MMD,JAN}. 

\begin{figure}[t]
    \centering
    \includegraphics[width= 0.99\columnwidth]{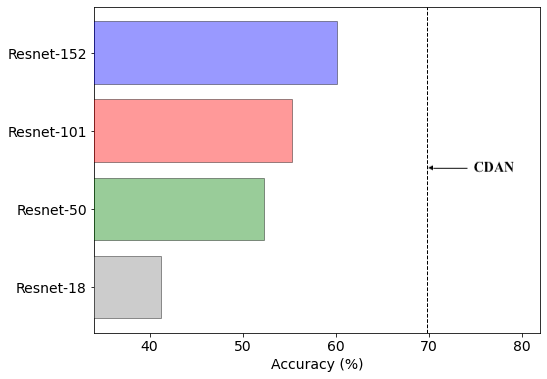}
    \caption{\textbf{Impact of powerful feature learners}. ``Source-only" performance of different architectures on the VISDA dataset~\cite{visda} compared to a sophisticated UDA approach CDAN~\cite{CDANLong}. The performance gap between the ``Source-only" classifiers and UDA steadily decreases with an increase in capacity of networks.}
    \label{fig:intro}
\end{figure}

As these distribution alignment strategies continue to mature, more sophisticated deep feature learners have also become available. For example, Figure~\ref{fig:intro} shows the native performance (on the target dataset) of different ResNet architectures (trained only using source data) for VisDA, a widely adopted UDA benchmark dataset. We also include the performance of a state-of-the-art adversarial UDA approach (CDAN~\cite{CDANLong}) for comparison. A striking observation is that the gap between target performance of the ``{source-only}'' model and that of the UDA algorithm steadily decreases with increasing number of convolution layers. This observation motivates us to revisit the use of subspace methods for UDA, in lieu of advanced alignment strategies, and to design a new subspace-based DA approach for improved generalization.

To this end, we propose a novel, subspace-based UDA approach that uses a pre-trained feature extractor based on sophisticated network architectures and a primary-auxiliary training algorithm to effectively trade-off the alignment and model fidelity objectives. More specifically, our approach is comprised of two key steps: (i) pre-training a feature extractor that utilizes labeled source data as well as unlabeled target data; (ii) solving the bi-level optimization of classifier training (posed as a \textit{primary} task) and subspace-alignment (posed as an \textit{auxiliary} task), wherein the feature extractor is not updated, unlike most existing UDA approaches. In this study, for the first time, we show that subspace alignment can provide comparable or sometimes even improved performance over state-of-the-art methods that use sophisticated alignment strategies (e.g., adversarial training), employ advanced regularizers (e.g., cyclical consistency) and are computationally more intensive.

Our findings, based on several widely adopted benchmarks in visual object recognition, show that global subspace alignment is a highly flexible and competitive baseline for UDA. The benefits of adopting a subspace alignment strategy, along with pre-trained feature extractors, for UDA are threefold: (i) In addition to providing computational efficiency, simpler alignment strategies can reduce the data requirements in the target domain; (ii) isolating feature learning and alignment steps eliminates the need for re-training the feature extractor from scratch for every new target domain, thereby enabling progressive adaptation; and finally, (iii) subspace alignment leads to intrinsically regularized models that we find to generalize better to challenging settings such as partial domain adaption, \textit{i.e.}, target domain contains only a subset of the classes observed in the source domain.

\subsection{Summary of Contributions}
Our contributions and key findings from this study are summarized below:
\begin{itemize}[noitemsep]
      \item A Target-Aware Feature Extractor (TAFE) pre-trained using both unlabeled target data and labeled source data;
       \item A novel primary-auxiliary formulation for UDA that performs subspace-alignment~\cite{Subspace_alignment} on features from a \textit{pre-trained} TAFE to achieve domain invariance;
       \item By posing subspace alignment as an auxiliary task for the primary task of obtaining well-calibrated classifiers, we are able to entirely dispense the need for adversarial learning, consistency-enforcing regularizers, and other extensive hyper-parameter choices;
       \item Our approach achieves higher or similar UDA performance, when compared to state-of-the-art approaches on several benchmarks;
      \item We find that, our approach is effective even with limited data in the target domain and is computationally efficient due to significant reduction in the number of parameters;
      \item Subspace-based DA leads to intrinsically well-regularized models that produce improved generalization even in partial DA settings;
      \item We show that, one can progressively adapt to additional target domains at test-time by recomputing only the subspace alignment and classifier parameters without the need to perform UDA from scratch.
      
\end{itemize}

\section{Related work}

\begin{figure*}[t]
	\centering
	\centerline{\includegraphics[width=1\linewidth]{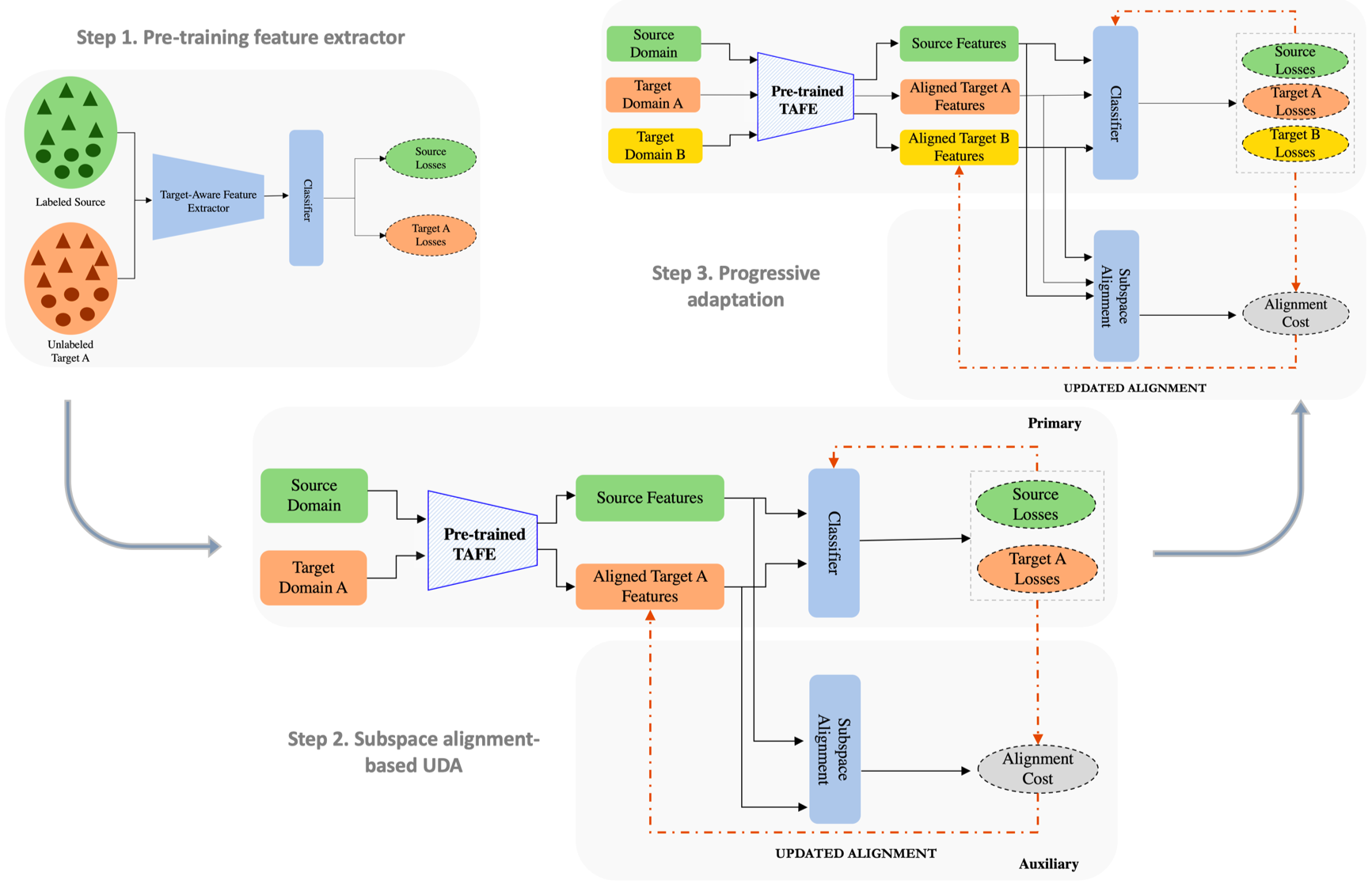}}
    \caption{\textbf{An overview of the proposed approach for subspace alignment-based UDA}. \textit{Step 1}: We train a Target-Aware Feature Extractor (TAFE) with a joint objective of decreasing empirical risk on labeled source and conditional entropy on the unlabeled target domain A; \textit{Step 2}: We leverage gradients from the \textit{primary task} of designing a well-calibrated classifier to guide the subspace alignment process, which is posed as an \textit{auxiliary task}. Note, in contrast to most existing UDA methods, the feature extractor is frozen after the pre-training phase; \textit{Step 3}: By isolating feature learning and domain alignment steps, our approach enables progressive adaptation to additional target domains at test-time, without requiring retraining from scratch. Note that, here only the classifier and the subspace alignment parameters are recomputed with unlabeled data from the target domain B.}
	\label{fig:Proposed approach}
\end{figure*}

\label{relatedwork}

\noindent \textbf{Unsupervised domain adaptation}:
Unsupervised domain adaptation has been an important problem of research in multiple application areas and a wide variety of solutions have been developed. Earlier works focused on adapting the features of source and target domains by minimizing statistical divergence between them \cite{Saenko2010AdaptingVC,Gong2012GeodesicFK,pan2010domain,sun2017correlation,Subspace_alignment,sun2015subspace}. These works can be analyzed through the foundational work of Ben David \etal ~\cite{Ben-David:2010:TLD:1745449.1745461}, which provides an upper bound on target error, $\epsilon(\mathcal{D}_T; h)$ on target data $\mathcal{D}_T$, that can be achieved using a hypothesis $h$ as the sum of three terms: 
\begin{equation}
\epsilon(\mathcal{D}_T; h) \leq \mathcal{L}(\mathcal{D}_S; h)+  \mathcal{L}_{\mathcal{H}} (\mathcal{D}_S, \mathcal{D}_T)+ \mathcal{L}_{\delta}(h),
\label{eq:error}
\end{equation}
where, the first term denotes the error in the source domain $\mathcal{S}$, the second term is the discrepancy between the source-target pair ($\mathcal{H}$-divergence), and the third term measures the optimal error achievable in both the domains (often assumed to be negligible). Under this context, there are two broad categories of methods -- ones that assume there exists a single hypothesis $h$ that can perform well in both domains (\textit{conservative}), and those that do not make that assumption (\textit{non-conservative})~\cite{shu2018a}. Successful state-of-the art methods use powerful feature extractors such as convolutional neural networks (CNNs), and aim to jointly minimize source error along with domain divergence error. Adversarial learning~\cite{cite:NIPS14GAN} has been the workhorse in these solutions, and can be implemented with different additional regularizers~\cite{cite:JMLR16RevGrad,CDANLong,CYCADA, UNIT,cite:CVPR17pix2pix}. Additionally some recent works have considered the use of pseudo-labeling and self-training~\cite{saito2017asymmetric,zou2019confidence,shot}, on top of these regularization strategies to further boost the UDA performance. While this work focuses on investigating the efficacy of subspace alignment in matching complex feature distributions, it is straightforward to extend our approach with additional regularizers or self-training protocols.

\noindent \textbf{Subspace-based adaptation}:
The key idea behind this class of methods is to represent source and target data distributions on lower-dimensional subspaces, align the subspaces, and subsequently project the target data onto the aligned subspace. A classifier is finally trained on the newly computed lower dimensional source data and evaluated on target data.  Popular approaches include~\cite{Gong2012GeodesicFK,Gopalan2011DomainAF,Subspace_alignment,sun2015subspace}. Geodesic-based methods \cite{Gopalan2011DomainAF,Gong2012GeodesicFK} compute a path along the manifold of subspaces (Grassmannian), and either project the
source and target onto points along that path~\cite{Gopalan2011DomainAF} or compute a linear map that projects source samples directly onto the target subspace~\cite{Gong2012GeodesicFK}. 
Furthermore, works such as \cite{Subspace_alignment,sun2015subspace} align the source and target subspaces using Procrustes methods \cite{Subspace_alignment}, or by considering distributional statistics along with subspace basis \cite{sun2015subspace}.

\noindent \textbf{Meta auxiliary learning}:
Meta-learning has been a recently successful approach in generalizing knowledge across related tasks~\cite{maml}. Broadly, meta-learning techniques can be grouped into three categories~\cite{maml} -- metric-based~\cite{koch2015siamese,vinyals2016matching}, model-based~\cite{Santoro:2016:MMN:3045390.3045585,MunkhdalaiY17} and optimization-based~\cite{maml,RaviL17}. Auxiliary learning on the other hand essentially focuses on increasing the performance of a primary task through the help of another related auxiliary task(s). This methodology has been applied to areas such as speech recognition~\cite{ToshniwalTLL17}, depth estimation, semantic segmentation~\cite{Liebel}, and reinforcement learning~\cite{JaderbergMCSLSK17}. The work closely related to ours is \textit{meta-auxiliary learning} \cite{metaauxillary}, which aims to improve $m$-class image classification performance (primary task) by solving a $k$-class classification problem (auxiliary task). This is done by establishing a functional relationship between the classes. In contrast, we formulate subspace-based domain alignment as the auxiliary to the primary task of building a classifier that works well in both source and target domains.

\section{Methods}
\label{methods}

\subsection{Overview}
The core idea behind our approach is to reduce the complexity of the alignment strategy by utilizing powerful deep feature extractors. To this end, we explore the use of simple subspace alignment to match source and target feature distributions. As illustrated in Figure \ref{fig:Proposed approach}, our UDA approach is comprised of two phases - The first phase involves pre-training a feature extractor using both labeled source and unlabeled target data. Once trained, the feature extractor is \textit{frozen} and the inferred source/target features are utilized for performing UDA. The second phase involves an alternating optimization between a classifier (primary task) and explicit domain alignment (auxiliary task) networks. The primary network updates the classifier given the source and the current best estimate of source-aligned target features, such that the inferred model is effective for both source and target domains. The auxiliary network, on the other hand, solves for subspace-based domain alignment, while minimizing both the alignment cost, and the loss from the primary network. Note, though the resulting alignment might be sub-optimal in terms of the pure alignment cost, it is systematically adapted to eventually improve the primary task.\\

\noindent \textbf{Notation:} In our setup, we assume access to labeled data from source domain $\{(x_s,y_s)\} \in \mathcal{D}_s$ and unlabeled data from the target domains $\{(x_t)\} \in \mathcal{D}_t$. 
We use the matrices $\mathrm{Z}_s, \mathrm{Z}_t$ to refer to the latent features for source and target data from the target-aware feature extractor (TAFE) $\mathcal{F}_{\Omega}$. $\mathrm{W}_s,\mathrm{W}_t $ denote the basis vectors of subspaces inferred from the latent features of source and target data $\mathrm{Z}_s, \mathrm{Z}_t$ respectively. 

\subsection{Target-Aware Feature Extractor}
\label{sec:Tafe}
As discussed earlier, we begin by constructing a Target-Aware Feature Extractor (TAFE) that utilizes both labeled source and unlabeled target data. In particular, we consider the joint objective of minimizing the empirical risk on labeled source data and conditional-entropy~\cite{shu2018a} on the unlabeled target data. By adopting a sophisticated feature learner and an appropriate optimization strategy, our goal is to improve the effectiveness of subspace-based alignment in UDA. 

Formally, denoting TAFE as $\mathcal{F}_\Omega$, our goal is to learn its parameters $\Omega$ along with the parameters $\Psi$ of a classifier $\mathcal{G}_\Psi$. 
For simplicity, we use the notation $\mathcal{P}_\Theta \coloneqq \mathcal{F}_\Omega \circ \mathcal{G}_\Psi$, where $\Theta \coloneqq \Omega \cup \Psi$. The losses used for the optimization include: (i) categorical cross-entropy loss, 
\begin{equation}
\label{eqn:source_ce}
    \mathcal{L}_{y}\left(\Theta ; \mathcal{D}_{s}\right)=\mathbb{E}_{x, y \sim \mathcal{D}_{s}}\left[y^{\top} \ln \mathcal{P}_{\Theta} (x)\right],
\end{equation}
for the labeled source data, (ii) conditional entropy~\cite{shu2018a} of the softmax predictions from the target data 
\begin{equation}
\label{eqn:target_ce}
 \mathcal{L}_{c}\left(\Theta ; \mathcal{D}_{t}\right)=-\mathbb{E}_{x \sim \mathcal{D}_{t}}\left[\mathcal{P}_{\Theta}(x)^{\top} \ln \mathcal{P}_{\Theta}(x)\right],
\end{equation}
 and (iii) class-balance loss~\cite{FrenchMF18} for the unlabeled target domain $\mathcal{L}_{cb}(\Theta ; \mathcal{D}_{t})$, which is implemented as binary cross-entropy between the mean prediction from the network over a mini-batch to that of a uniform distribution -- this loss regularizes network behavior when the data exhibits large class imbalance. The overall loss function can thus be defined as
\begin{equation}
\label{eqn:sum losses}
    \mathcal{L}_{\mathcal{P}}(\Theta)=\mathcal{L}_{y}\left(\Theta ; \mathcal{D}_{s}\right)+\lambda_c \mathcal{L}_{c}\left(\Theta ; \mathcal{D}_{t}\right)+\lambda_{cb}\mathcal{L}_{cb}(\Theta ; \mathcal{D}_{t}). 
\end{equation}Once trained, the parameters of the pre-trained TAFE model are frozen, and the latent features $\mathrm{Z}_s, \mathrm{Z}_t$ are extracted \textit{i.e.}, $\mathrm{Z}_s=\mathcal{F}_{\Omega}(x_s)~\forall~x_s\in\mathcal{D}_{s}$ and $\mathrm{Z}_t=\mathcal{F}_{\Omega}(x_t)~\forall~x_t\in\mathcal{D}_{t}$.
The classifier layer $\mathcal{G}_\Psi$, however, is further refined in the second phase of our algorithm. We detail TAFE training in Algorithm~\ref{algo:algo_tafe}.
\RestyleAlgo{boxruled}
\begin{algorithm}[t]
	
	\KwIn{labeled source and unlabeled target datasets $\{\mathrm{X}_s,\mathrm{y}_s\} $ and $\{\mathrm{X}_t\}$ from source and target domains $\mathcal{D}_S $ and  $\mathcal{D}_T$}
	\KwOut{Latent features of source and target data $\mathrm{Z}_s,\mathrm{Z}_t$}
	\textbf{Initialize}: Feature extractor $\mathcal{F}_{\Omega}$ , classifier $\mathcal{G}_{\Psi}$ and combined network $\mathcal{P}_\Theta \coloneqq \mathcal{F}_\Omega \circ \mathcal{G}_\Psi$. \\
	Set hyper-parameters $\lambda_c,\lambda_{cb},N $

		\tcp{Train $\mathcal{P}_\Theta$}
		\For{epoch \textbf{in} $N$}{
		
        Compute source loss $\mathcal{L}_{y}$ using \eqref{eqn:source_ce}\\
        Compute target loss $\mathcal{L}_{c}$ using \eqref{eqn:target_ce}
        and $\mathcal{L}_{cb}$ following Sec~\ref{sec:Tafe}

        Update $\mathcal{P}_{\Theta}$ with Eq~\eqref{eqn:sum losses} 
		 }
	
        \tcp{Extract source and target features}
        $\mathrm{Z}_s=\mathcal{F}_{\Omega}(x_s)~\forall~x_s\in\mathcal{D}_{s}$ and\\ $\mathrm{Z}_t=\mathcal{F}_{\Omega}(x_t)~\forall~x_t\in\mathcal{D}_{t}$. 

	\caption{Proposed algorithm to train TAFE.}\label{algo:algo_tafe}
\end{algorithm}

\subsection{Proposed UDA Approach}
In this phase, we include an explicit distribution matching strategy based on linear subspace alignment of pre-trained TAFE features. To this end, we adopt a primary-auxiliary task formulation, wherein the primary task refines classifier $\mathcal{G}_{\Psi}$ based on source features and source-aligned target features, while the auxiliary task is aimed at aligning the source and target subspaces. 

\subsubsection{\textbf{\textit{Primary Task}  Classifier Update}}
\label{sec:primary}
In this task, our goal is to achieve effective class discrimination in both source and target domains. With inputs as latent features extracted from the pre-trained TAFE $\mathcal{F}_{\Omega}$, we fine-tune the parameters of the classifier $\mathcal{G}_{\Psi}$.
Denoting $z = \mathcal{F}_{\Omega}(x)$, we use the same losses as in Eq.~\eqref{eqn:sum losses} to train $\mathcal{G}_{\Psi}$ i.e., 
cross-entropy loss on source features\begin{equation*}
    \mathcal{L}_{y}\left(\Psi ; \mathcal{D}_{s}\right)=\mathbb{E}_{x, y \sim \mathcal{D}_{s}}\left[y^{\top} \ln \mathcal{G}_{\Psi} (z)\right], \text{ where }z = \mathcal{F}_{\Omega}(x),
\end{equation*} conditional-entropy loss on source-aligned target features\begin{equation*}
\mathcal{L}_{c}\left(\Psi ; \mathcal{D}_{t}\right)= -\mathbb{E}_{z \sim \mathcal{D}_{t}}\left[\mathcal{G}_{\Psi}(\mathcal{A}_{\Phi}(z))^{\top} \ln \mathcal{G}_{\Psi}(\mathcal{A}_{\Phi}(z))\right],\end{equation*}where $\mathcal{A}_{\Phi}$ is the alignment operation and the class-balance loss $\mathcal{L}_{cb}$ on the target features. 
Overall,
\begin{equation}
\label{eqn:classifier_loss}
    \mathcal{L}_{\mathcal{G}}(\Psi)=\mathcal{L}_{y}(\Psi; \mathcal{D}_s)+\gamma_c \mathcal{L}_{c}(\Psi; \mathcal{D}_t)+\gamma_{cb}\mathcal{L}_{cb}(\Psi; \mathcal{D}_t). 
\end{equation}
Note that, in the definition of $\mathcal{L}_{c}$, the target domain features are first transformed using the auxiliary network $\mathcal{A}_{\Phi}$ (defined in Section~\ref{sec:SA-network}) prior to applying the classifier. 

\subsubsection{\textbf{\textit{Auxiliary Task}  Domain alignment}}
\label{sec:SA-network}
In general, as a generative model for a dataset, a single linear subspace or even a union of linear subspaces is known to be insufficient. However, given powerful feature learners, we posit that, an alternating optimization between the primary task of learning a generalizable classifier and the auxiliary task of domain alignment can make even the simple subspace alignment~\cite{Subspace_alignment} highly effective in UDA. 

\noindent \textit{Closed-form subspace alignment:} Let us denote the basis vectors for the $d$-dimensional subspaces inferred from source and target domains using the matrices $\mathrm{W}_s$ and $\mathrm{W}_t$ respectively, and they satisfy 
$\mathrm{W}_s^T\mathrm{W}_s = \mathbb{I}$, $\mathrm{W}_t^T\mathrm{W}_t = \mathbb{I}$, where $\mathbb{I}$ denotes the identity matrix. 
The subspaces are inferred using singular value decomposition of source/target domain latent features $\mathrm{Z}_s, \mathrm{Z}_t$.
The alignment between two subspaces can be parameterized as an affine transformation ${\Phi}$, \ie
\begin{align}
    \label{eq:SAobjective}
    \begin{split}
    {\Phi}^* = \argmin_{\Phi} \norm{\mathrm{W}_t {\Phi} - \mathrm{W}_s}_F^2,
    \end{split}
\end{align}where, $\norm{.}_F$ denotes the Frobenius norm. The solution to this alignment cost \eqref{eq:SAobjective} can be obtained in closed-form \cite{Subspace_alignment} as
\begin{equation}
    \label{eq:globalsoln}
        {\Phi}^* = (\mathrm{W}_t)^{\top} \mathrm{W}_s.
\end{equation}This implies that the adjusted coordinate system, also referred as the \textit{source-aligned target subspace} can be constructed as 
 \begin{equation} \bar{\mathrm{W}}_{t} =  \mathrm{W}_t (\mathrm{W}_t)^{\top} \mathrm{W}_s.
    \label{eqn:align}
\end{equation}
Since the primary task invokes the classifier optimization using features in the ambient space, we need to re-project the target features using $\bar{\mathrm{W}}_t$, \textit{i.e.},
\begin{align}
{\hat{\mathrm{Z}}_t}^* &= \mathcal{A}_{\Phi}(\mathrm{Z}_t) = \argmin_{\hat{\mathrm{Z}}_t} \norm{\hat{\mathrm{Z}}_t \mathrm{W}_s -\hat{\mathrm{Z}}_{t} \bar{\mathrm{W}}_t}_{F}^2 \nonumber \\
&=\argmin_{\hat{\mathrm{Z}}_t} \norm{\hat{\mathrm{Z}}_t \mathrm{W}_s -\hat{\mathrm{Z}}_{t}  \mathrm{W}_t (\mathrm{W}_t)^{\top} \mathrm{W}_s}_{F}^2,
\end{align}where $\hat{\mathrm{Z}}_t^*$ denotes the modified target features. The solution to this optimization problem can be obtained in closed-form as
\begin{align}
    \label{eq:reproj}
    \mathcal{A}_{\Phi}(\mathrm{Z}_t) &= \mathrm{Z}_t\mathrm{W}_t{\Phi}^*\mathrm{W}_s^{\top},
\end{align}where ${\Phi}^*$ is computed using \eqref{eq:globalsoln}.

 \RestyleAlgo{boxruled}
\begin{algorithm}[t]

\caption{Proposed algorithm for unsupervised visual domain adaptation}
	\label{algo:algo}
	\KwIn{
	Labeled source features $\{\mathrm{Z}_s,\mathrm{y}_s\} $ and 
	unlabeled target features $\{\mathrm{Z}_t\}$ from TAFE $\mathcal{F}_{\Omega}$.
	Source and target subspaces $\mathrm{W}_s$, $\mathrm{W}_t$ }
	
	\textbf{Initialize}: $\mathcal{G}_{\Psi}$ from $\mathcal{F}_{\Omega}$, $\Phi$ using \eqref{eq:globalsoln}; Hyper-parameters $\gamma_c,\gamma_{cb}, n_{iter},T_1,T_2 $.

\textbf{Training Phase}:\newline  
     Split: $\mathrm{Z}_S^{\dagger} , \mathrm{Z}_S^{\ddagger} \leftarrow  \mathrm{Z}_S$ and $\mathrm{Z}_T^{\dagger} , \mathrm{Z}_T^{\ddagger} \leftarrow \mathrm{Z}_T$ \newline  
 	\For{iter \textbf{in} $n_{iter}$}{
 	     \tcp{ update $\mathcal{G}_{\Psi}$}
 	    \For{$t_1$ \textbf{in} $T_1$}{
           		   
			Compute $\bm\hat{\mathrm{Z}}_t^{\dagger} = \mathrm{Z}_t^{\dagger}\mathrm{W}_t\mathrm{\Phi}^*\mathrm{W}_s^T $ following \eqref{eq:reproj}\;
			$\bm\hat{\mathrm{y}}_s^{\dagger} = \mathcal{G}_{\Psi}(\bm\hat{\mathrm{Z}}_s^{\dagger})$\; 
			$\bm\hat{\mathrm{y}}_t^{\dagger} = \mathcal{G}_{\Psi}(\bm\hat{\mathrm{Z}}_t^{\dagger})$\;

			Compute $\mathrm{L}_{\mathcal{G}}$ using \eqref{eqn:classifier_loss}\;
		
			Update $\Psi^* = \arg \min_{\Psi} \mathrm{L}_{\mathcal{G}} $ \;}
            
	        \tcp{ update $\mathcal{A}_{\Phi}$}
			\For{$t_2$ \textbf{in} $T_2$}{

			Compute $\bm\hat{\mathrm{Z}}_t^{\ddagger} $ using \eqref{eq:reproj}\; 
			Compute $\bm\hat{\mathrm{y}}_t^{\ddagger} =\mathcal{G}_{\Psi^*}(\bm\hat{\mathrm{Z}}_t^{\ddagger})  $; 

			Compute $\mathcal{L}_{\mathcal{A}}$ using \eqref{eqn:loss_A}\; 
			Update $\mathrm{\Phi}^* = \arg \min_{\Phi} \mathrm{L}_{\mathcal{A}_{\Phi}} $ \;
			}
 		}

\end{algorithm}

Note that though we develop our formulation by aligning the target subspace onto the source, one can equivalently project the source subspace onto the target.
\noindent \textit{Task-dependent tuning of subspace alignment: }
Since the overall objective is to refine the auxiliary network parameters to maximally support the primary task, we propose to include the terms $\mathcal{L}_c$ and $\mathcal{L}_{cb}$ from \eqref{eqn:classifier_loss} to the alignment cost in \eqref{eq:SAobjective}, 
\begin{equation}
\mathcal{L}_{\mathcal{A}}(\Phi) = \norm{\mathrm{W}_t {\Phi} - \mathrm{W}_s}_F^2 + \gamma_c \mathcal{L}_{c}(\Phi;\mathcal{D}_t)+\gamma_{cb}\mathcal{L}_{cb}(\Phi;\mathcal{D}_t).
    \label{eqn:loss_A}
\end{equation}
Note that, when we make this modification, there no longer exists a closed-form solution. Hence, we adopt an approach that takes in gradients from the primary task to adjust ${\Phi}$. To enable this end-to-end training of both the primary and auxiliary tasks, we implement \textit{subspace alignment} as a network $\mathcal{A}$ that parameterizes ${\Phi}$ as a fully connected layer of $d$ neurons without any non-linear activation function or bias.

\begin{table*}[h]
    \begin{center}
    \begingroup
 \renewcommand*{\arraystretch}{1.3}
    \begin{tabular}{c|cccccc|c}
    
    \hline

 \textbf{Method} & \textbf{I $\rightarrow$ P} & \textbf{P $\rightarrow$ I} & \textbf{I $\rightarrow$ C} & \textbf{C $\rightarrow$ I} & \textbf{C $\rightarrow$ P} & \textbf{P $\rightarrow$ C} & \textbf{Average} \\ 
\hline
No Adaptation & 76.5 & 88.2 & 93 & 84.3 & 69.1 & 91.2 & 83.7 \\
DAN~\cite{DAN} & 74.5$\pm$0.4 &82.2$\pm$0.2 & 92.8$\pm$0.2 & 86.3$\pm$0.4 & 69.2$\pm$0.4 & 89.8$\pm$0.4 & 82.5 \\
DANN~\cite{cite:JMLR16RevGrad} & 75.0$\pm$0.6 & 86.0$\pm$0.3 & 96.2$\pm$0.4   & 87.0$\pm$0.5 &74.3$\pm$0.5 &91.5$\pm$0.6 &85.0 \\
JAN~\cite{JAN} & 76.8$\pm$0.4 &88.0$\pm$0.2& 94.7$\pm$0.2& 89.5$\pm$0.3& 74.2$\pm$0.3& 91.7$\pm$0.3& 85.8 \\

CDAN+E~\cite{CDANLong} & {\bf \em 77.7$\pm$0.3 }& {\bf \em 90.7$\pm$0.2} & \textbf{97.7$\pm$0.3} & \textbf{91.3$\pm$0.3} & {\bf \em 74.2$\pm$0.2} & {\bf \em 94.3$\pm$0.3} & {\bf \em 87.7} \\
\hline
Ours & \textbf{80.16$\pm$0.2} & \textbf{95.5$\pm$0.3} & {\bf \em 97.3$\pm$0.4} & {\bf \em 90.9$\pm$0.3} & \textbf{79.3$\pm$0.2} & \textbf{97$\pm$0.6} & \textbf{90.02} \\

    \hline
\end{tabular}
  
    \endgroup
    \caption{\small{\textbf{UDA performance on the ImageCLEF dataset}. Best performance is shown in {\bf bold}, and the second best in {\em \bf \em bold italic}.}}
    \label{table:image_clef}
    \end{center}
    \end{table*}

\noindent \textbf{Primary-Auxiliary Objective.} The overall objective of our primary-auxiliary learning for UDA can be formally written as the following bi-level optimization:
\begin{align}
&\min_{\Psi} \mathcal{L}_{\mathcal{G}}\left(\Psi; \mathrm{Z}_s,\mathrm{y}_s, \mathcal{A}_{\Phi^*}(\mathrm{Z}_t)\right), \\
\nonumber \text{where,} \quad {\Phi}^* = &\arg \min_{{\Phi}} \mathcal{L}_{\mathcal{A}}\bigg({\Phi}; \mathrm{W}_s,\mathrm{W}_t,  \mathcal{P}_{\Theta}(\mathcal{A}_{\Phi}(\mathrm{Z}_t))\bigg).
\label{eqn:obj}
\end{align}We now describe the algorithm for solving this objective. 


\subsection{Algorithm}
Given the primary and auxiliary task formulations, one can adopt different training strategies to combine their estimates: (i) \textit{Independent}: This is the classical approach, where the alignment obtained by solving \eqref{eq:SAobjective} is used to infer the classifier parameters; (ii) \textit{Joint}: This jointly optimizes for both networks together, similar to existing domain adaptation methods; (iii) \textit{Alternating}: This alternating style of optimization solves for the primary task with the current estimate of the alignment, and subsequently updates the auxiliary network with both primary and auxiliary losses. As we will show later, that this alternating optimization strategy works the best in comparison to the other two. We now describe the alternating optimization strategy.

\noindent \textbf{Initialization phase}: The choice of initial states for the parameters of both the primary and auxiliary networks is crucial to the performance of our algorithm. First, we pre-train TAFE $\mathcal{F}_{\Omega}$ using the loss function \eqref{eqn:sum losses} without any explicit domain alignment. 
We then fit $d$-dimensional subspaces, $\mathrm{W}_s$ and $\mathrm{W}_t$, to the features obtained using $\mathcal{F}$ for both the source and target domains. 
Note, the feature extractor is not updated for the rest of the training process, and hence the subspace estimates are fixed. 
The initial state of ${\Phi}$, \ie~alignment matrix between the two subspaces, is obtained using \eqref{eq:globalsoln} and we fine-tune the classifier $\mathcal{G}_\Psi$.

\noindent \textbf{Training phase}: In order to enable information flow between the two tasks, we propose to allow the auxiliary task to utilize gradients from the primary task. Similarly, the estimated alignment is applied to the target data while updating the classifier parameters in the primary task.
The two tasks are solved alternatively until convergence -- during the auxiliary task optimization, we freeze the classifier parameters and update $\mathcal{A}_{\Phi}$. Since the feature extractor $\mathcal{F}_{\Omega}$ is fixed, there is no need to recompute the subspaces. In our implementation, we find that optimizing the auxiliary task using a held-out validation set, distinct from that used for the primary task, helps in the convergence. Given the estimate for ${\Phi}$, we freeze the auxiliary network $\mathcal{A}_{\Phi}$ and update the classifier network using source features and source-aligned target features to minimize the primary loss in \eqref{eqn:classifier_loss}. Upon convergence (typically within $5-10$ iterations on all datasets considered), optimal values for both $\mathcal{A}_{\Phi}$ and $\mathcal{G}_{\Psi}$ are returned. A detailed listing of this process is provided in algorithm~\ref{algo:algo}.
\subsection{Progressive Domain Adaptation}
\label{sec:progressive DA}
One key design choice of our approach is to isolate feature learning and explicit alignment steps.  
This design choice enables the use of our approach for progressive domain adaptation to additional target domains at test-time. Existing approaches that aim to achieve invariance between the source and target distributions typically need to be retrained from scratch for every new target domain.

Specifically, consider the vanilla UDA setting, where we initially have access to labeled data from source domain $\mathcal{D}_s$ and unlabeled data from target domain $\mathcal{D}_A$. We first adapt the model $\mathcal{P}_{\Theta}$ from $\mathcal{D}_s \rightarrow \mathcal{D}_A$ and deploy the model. 
We now assume that a new unlabeled data from another domain $\mathcal{D}_{B}$ is collected and the model needs to be adapted to this new domain at test-time (\textit{Step 3} in Figure \ref{fig:Proposed approach}). 
To achieve this, we re-purpose the TAFE model inferred for performing UDA between $\mathcal{D}_s$ and $\mathcal{D}_A$, and compute the features $\mathrm{Z}_s,\mathrm{Z}_A, \mathrm{Z}_{B}$ for data $\mathrm{X}_s,\mathrm{X}_A, \mathrm{X}_{B}$ respectively. We also transform $\mathrm{Z}_A$ to $\hat{\mathrm{Z}}_A^*$ using the optimal transformation matrix $A_{\Phi^{*}}$ inferred using our algorithm.

For progressive adaptation, we need to compute the alignment matrix $A_{\Upsilon^{*}}$ for matching the feature distributions from the new target $\mathrm{Z}_B$ and the collection of two observed domains, namely $\mathrm{Z}_s$ and $\hat{\mathrm{Z}}_A^*$. To this end, we repeat Algorithm \ref{algo:algo}, where we use the collection of labeled source data $(\mathrm{Z}_s, \mathrm{y}_s)$ and the pseudo-labeled data from target domain A $(\hat{\mathrm{Z}}_A^*, \mathrm{y}_A)$ as the source data for performing UDA to target domain B. Note that, the pseudo-labels $\mathrm{y}_A$ are obtained using our predictions for target domain A. By avoiding the need to retrain the feature extractor from scratch and leveraging the data from other target domains observed so far, our approach offers a flexibility that is not found in any of the state-of-the-art UDA methdods.

\section{Results and Findings}
\label{experiments}
We evaluated the proposed method on four widely used visual domain adaptation tasks -- digits, ImageCLEF, VisDA-2017 challenge, and Office-Home datasets, and present comparisons to several state-of-the-art domain adaptation techniques. Across all the experiments, an 80-20 random split of source and target training data was performed to update the primary and auxiliary tasks.
All experiments were repeated thrice and we report the mean and standard deviation for each case. We implemented all our algorithms and performed empirical studies using the PyTorch framework ~\cite{paszke2017automatic}.

\subsection{ImageCLEF-DA}
\label{sec-imageclef}
\noindent \textbf{Dataset}: ImageCLEF\footnote{\url{http://imageclef.org/2014/adaptation}} is organized by selecting common categories of images shared by three public image datasets (domains):  \textit{ImageNet ILSVRC 2012} (\textbf{I}), \textit{Caltech-256} (\textbf{C}), and \textit{Pascal VOC 2012} (\textbf{P}). There are $12$ categories, with $50$ images each,  resulting in a total of $600$ images in each domain. We conduct $6$ experiments by permuting the $3$ domains : \textbf{I} $\rightarrow$ \textbf{P}, \textbf{P} $\rightarrow$ \textbf{I}, \textbf{I} $\rightarrow$ \textbf{C}, \textbf{C} $\rightarrow$ \textbf{I}, \textbf{C} $\rightarrow$ \textbf{P}, \textbf{P} $\rightarrow$ \textbf{C}.

\begin{figure*}[t!]
    \centering
    \subfloat[Ablation study]{{\includegraphics[width=0.4\linewidth]{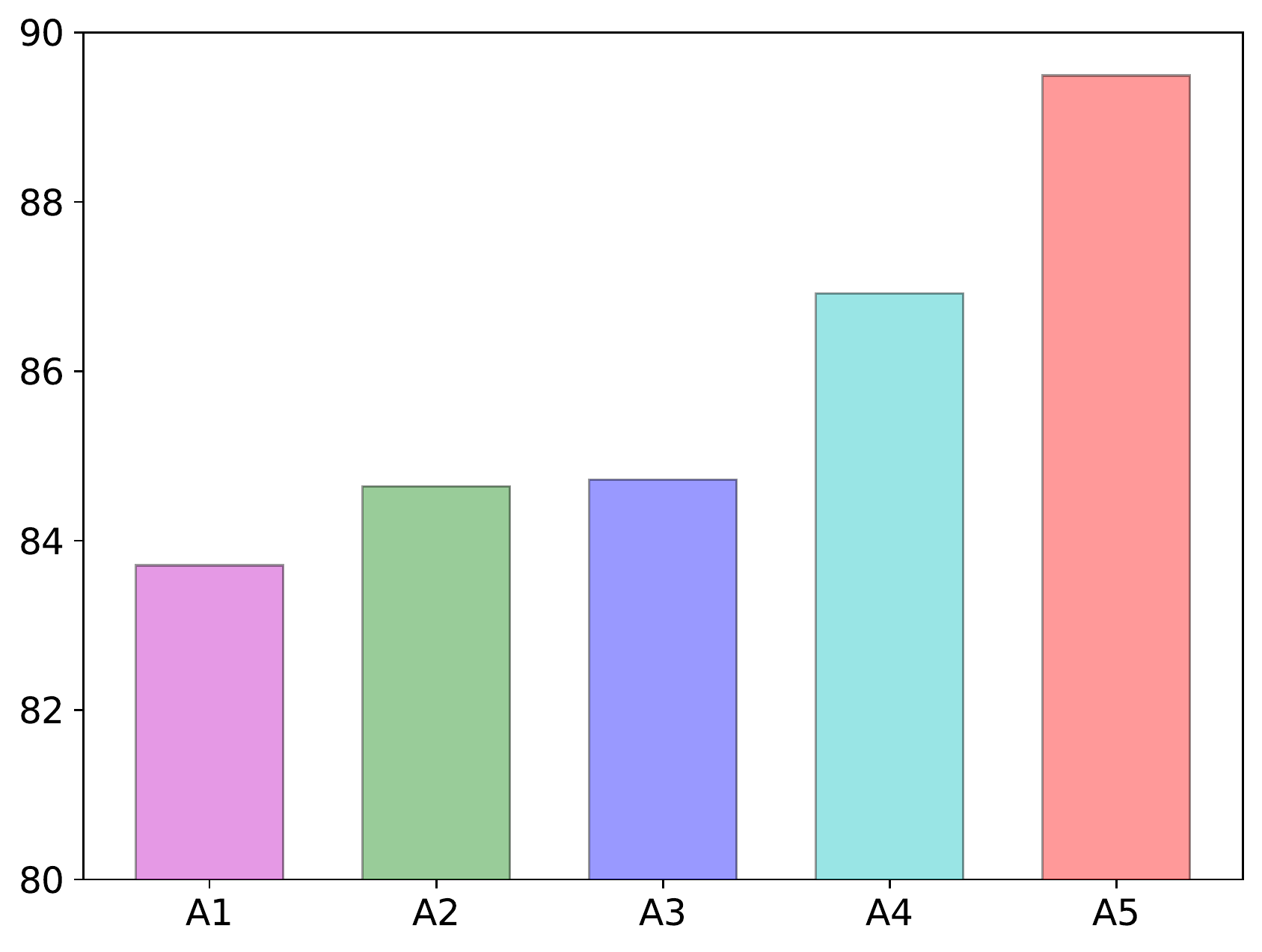} }}%
    \hfill
    \subfloat[Dynamics of ${\Phi}$ across iterations]{{\includegraphics[width=0.57\linewidth]{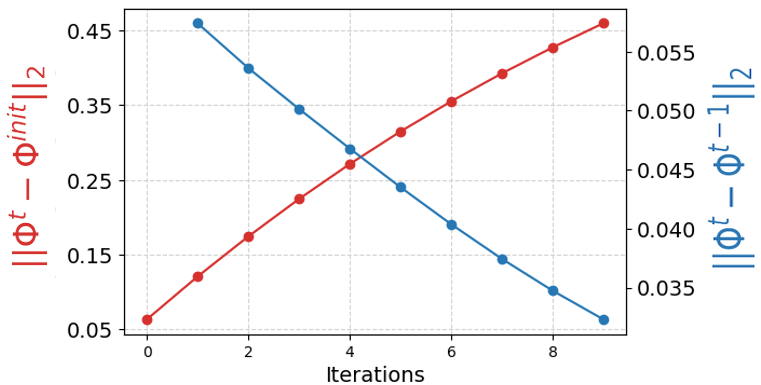} }}%
    \caption{\textbf{Behavior of the proposed UDA algorithm}. (a) Ablating different components in the proposed method against adaptation performance on the ImageCLEF dataset. See text in Section \ref{sec-imageclef} for notation.
    (b) Changes in ${\Phi}$ from  $\mathrm{\Phi^{init}}$ \eqref{eq:globalsoln}~ across iterations are represented by the red Curve while the blue curve denotes successive differences in ${\Phi}$}%
    \label{fig:ablation}%
\end{figure*}

\noindent \textbf{Model}: Our TAFE model is initialized using the pre-trained ResNet-50 architecture~\cite{ResNet,russakovsky2015imagenet} and is fine-tuned following the objective in \eqref{eqn:sum losses} with $\lambda_c$ and $\lambda_{cb}$ set at $0.1$. We then obtain the $2048$-dimensional latent features from the penultimate layer of TAFE. The source and target subspaces of dimension $800$ are constructed from these features using SVD. The classifier network is chosen to be the last fully connected layer, subsequently refined using the SGD optimizer with learning rate $1e-4$ and momentum $0.9$. The subspace alignment network, which is a single linear layer, is trained with learning rate of $1e-3$ using the Adam optimizer~\cite{ADAM}. 
The proposed approach is compared against a number of baseline methods including \cite{CDANLong,DAN,cite:JMLR16RevGrad,JAN} and the results are reported in Table \ref{table:image_clef}. The results clearly show that our approach based on subspace alignment improves performance by nearly 3 percentage points over sophisticated adversarial learning methods.

\noindent \textbf{Ablation Study}: In order to understand the impact of different components in our approach, we perform an ablation study on this dataset. We describe each setting in this experiment next: 
\begin{itemize}[noitemsep]
    \item[$\mathbf{A_1}$]  \emph{No Adaptation}: A baseline method where we use the classifier trained using the source data directly on the target features without any adaptation. 
    \item[$\mathbf{A_2}$]  \emph{Primary Only}: We leave out the auxiliary task, but include all the losses used in the primary task described in equation \eqref{eqn:sum losses}, with $\mathcal{A}_{\Phi} = \mathbb{I}$.
    \item[ $\mathbf{A_3}$] \emph{Independent}: Here, we use the closed form solution in subspace alignment from equation \eqref{eqn:align}, and then solve for the primary task in \eqref{eqn:sum losses} independently. 
    \item[ $\mathbf{A_4}$] \emph{Joint Optimization}: We employ a joint optimization strategy, wherein we jointly update the alignment ${\Phi}$, and the classifier together. 
    \item[$\mathbf{A_5}$]  \emph{Alternating Optimization}: This is our proposed strategy that updates ${\Phi}$ and the classifier in an alternating fashion. 
\end{itemize}The results from the study are illustrated in Figure \ref{fig:ablation}(a). A key observation is that, since the alignment strategy is weak, when done independently it does not lead to any performance gains. However, the proposed optimization provides a significant improvement over even a joint optimization strategy.

\begin{table}[t]

    \label{tab:VISDA_Digits}
    \begin{center}
    \vspace{0.1in}

    \subfloat[Digits datastets]{
    \renewcommand*{\arraystretch}{1.4}
    \begin{tabular}{c|p{0.5in}p{0.4in}p{0.4in}|c}
    
    \hline
    \textbf{Method}& \small{\textbf{MNIST$\rightarrow$ USPS}} & \small{\textbf{USPS$\rightarrow$ MNIST}} & \small{\textbf{SVHN$\rightarrow$ MNIST}}&\textbf{Avg} \\
    \hline
    No Adaptation & 94.8 & 49 & 60.7&68.2 \\
    DeepCoRAL~\cite{deepcoral} & 89.3& 91.5& 59.6&80.1 \\
    MMD~\cite{MMD} & 88.5& 73.5& 64.8&75.6\\
    DANN~\cite{cite:JMLR16RevGrad} & 95.7& 90.0& 70.8&85.5\\
    ADDA~\cite{tzeng2017adversarial} & 92.4 & 93.8 & 76.0&87.4\\
    DeepJdot~\cite{bhushan2018deepjdot} & 95.6 & 96.0 & \textbf{96.7}&\textbf{\textit{96.1}} \\
    CyCADA~\cite{CYCADA} & 95.6 & \mbox{\bf \em 96.5} & 90.9 & 94.3\\
    UNIT~\cite{UNIT} & \mbox{\bf \em 95.9} &93.5& 90.5&93.3 \\
    GenToAdapt~\cite{GTA} & 95.3& 90.8 &92.4 & 92.8\\
    \hline
    Ours & \textbf{96.2$\pm$0.3} & \textbf{97.4$\pm$0.4} & \mbox{\bf \em 95.6$\pm$0.2} & \textbf{96.4}\\
      \hline
 
    \end{tabular}
   \label{table:digits}
    }
    \vspace{0.1in}
    \subfloat[VISDA-2017]{
    \renewcommand*{\arraystretch}{1.3}
    \label{table:visda}
    \centering
     \begin{tabular}{c|c}
     \hline
        \textbf{Method}& \textbf{Average}  \\\hline
        No Adaptation & 54.2\\
         JAN~\cite{JAN} & 61.6 \\
        GTA~\cite{GTA} & 69.5 \\
        CDAN~\cite{CDANLong} & \mbox{\bf \em 70.2}\\
        \hline
        
       Ours &  \textbf{73.02$\pm$0.3}\\
        \hline
        \end{tabular}
          \label{table:visda}
    }
    \caption{\small{\textbf{Performance of the proposed method on Digits and VISDA-2017 datasets}. We highlight the best performing technique in {\bf bold}, and the second best in {\bf \em bold italic}.}}
    \end{center}
    \vspace{-0.2in}
    \end{table}

\noindent\textbf{Convergence of ${\Phi}$}:
Through Figure \ref{fig:ablation}(b) we report the training behavior of the alignment matrix ${\Phi}$ returned by the auxiliary network $\mathcal{A}_{\Phi}$. 
While the red curve in Figure \ref{fig:ablation}(b) indicates the change in ${\Phi}$ across iterations indexed by $t$ w.r.t the closed form solution ${\Phi}^{init}$ obtained in \eqref{eq:globalsoln}, the blue curve represents the successive difference in ${\Phi}$ across iterations. As expected, the estimate for ${\Phi}$ changes non-trivially from ${\Phi}^{init}$, eventually converging to a solution that leads to maximal classification performance. Note, in all our experiments, we find that the ${\Phi}$ returned by the auxiliary network is always a well-conditioned, full rank matrix. 



\subsection{Digits classification}

\noindent\textbf{Datasets:} We consider three data sources for the digits classification task: USPS \cite{hull1994database}, MNIST \cite{lecun2010mnist}, and the Street View House Numbers (SVHN) \cite{netzer2011reading} datasets.
Each of these datasets have 10 categories (digits from 0-9). 
We perform the following three experiments in this task.
a) MNIST $\rightarrow$ USPS, b) USPS $\rightarrow$ MNIST, and c) SVHN $\rightarrow$ MNIST and report the accuracies on the standard target test sets.


\noindent\textbf{Model:} The TAFE model used for all these $3$ tasks is based on the architecture from~\cite{bhushan2018deepjdot}, which is comprised of six $3 \times 3$ convolutional layers containing $\{32, 32, 64, 64, 128, 128\}$ filters with ReLU activations and two fully-connected layers of $128$ and $10$ (number of classes) hidden units. The Adam optimizer with learning rate $2e^{-4}$ was used to update the  model using a mini-batch size of $512$ for the two domains. We compare our results with a number of state-of-the-art methods and the results are shown in Table \ref{table:digits}. Our approach achieves the highest accuracy averaged across all three digits datasets, surpassing state-of-the-art in two out of three cases, and marginally below DeepJDOT \cite{bhushan2018deepjdot} in the case of SVHN $\rightarrow$ MNIST. 


\subsection{VisDA-2017}
\noindent\textbf{Dataset}:
VisDA-2017 is a challenging simulation-to-realworld dataset with two highly distinct domains: 
\textit{Synthetic}, renderings of 3D models from different angles and with different lightning conditions; \textit{Real} which are natural images. This dataset contains over 280K images across 12 classes.

\begin{figure*}[t]
    \centering
    \subfloat[No adaptation]{{\includegraphics[trim={5cm 5cm 5cm 5cm},clip,width=0.35\linewidth]{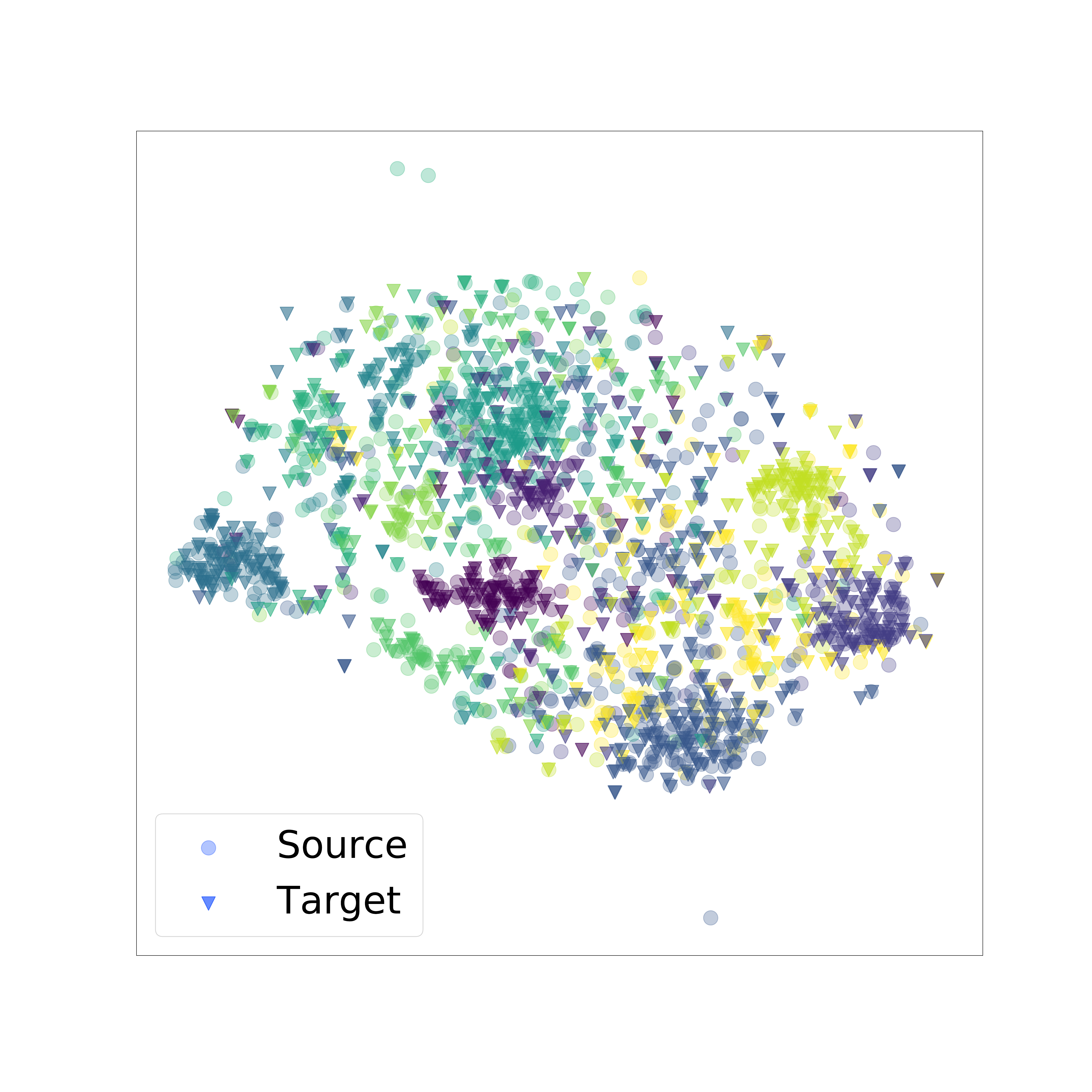} }}%
    \qquad
    \subfloat[After adaptation]{{\includegraphics[trim={5cm 5cm 5cm 5cm},clip,width=0.35\linewidth]{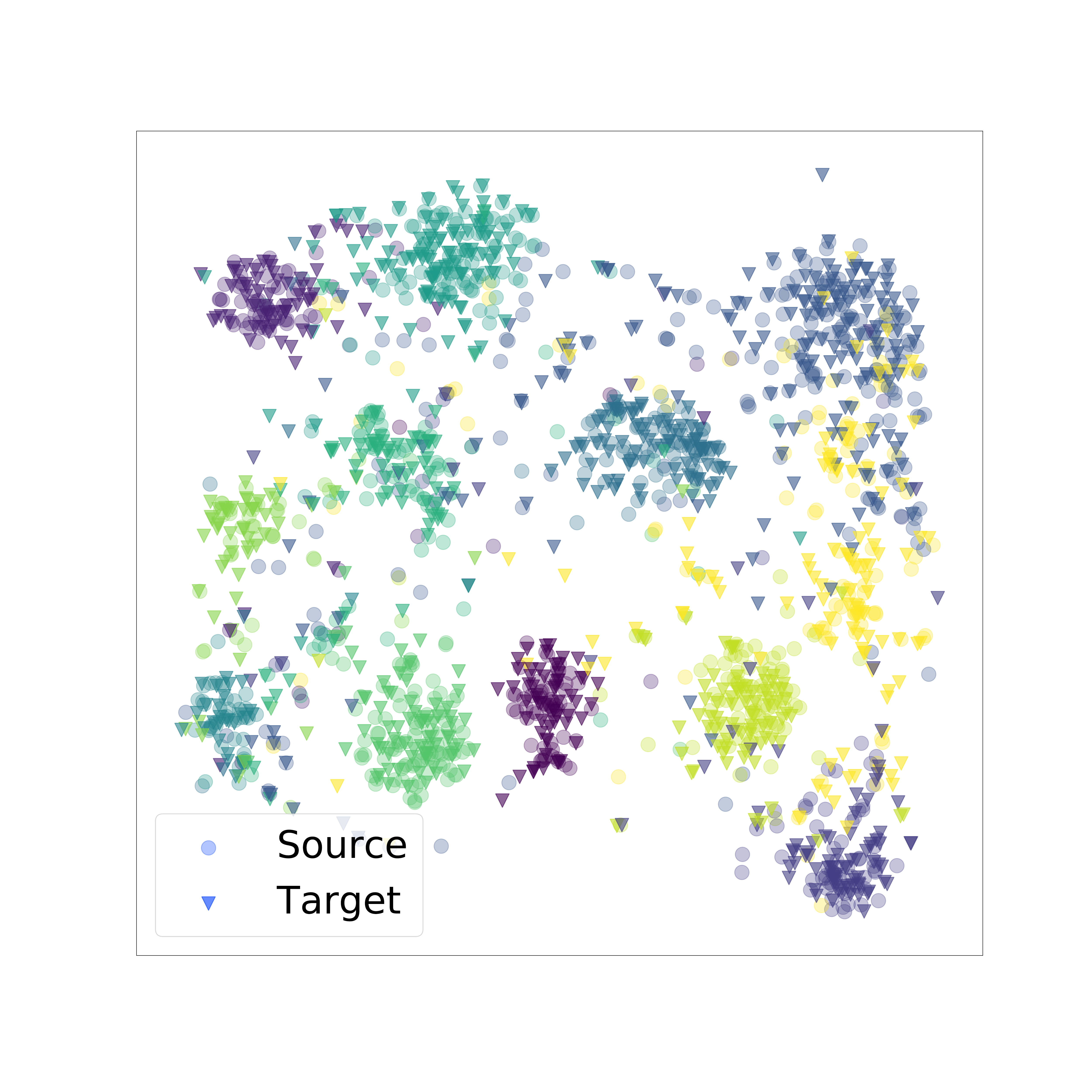} }}%
    \caption{\textbf{Comparing t-SNE embeddings before and after UDA on the {VisDA-2017} dataset}. We observe improved alignment between the class boundaries of the source and target domains.}%
    \label{fig:tsne}%
\end{figure*}



\begin{table*}[t]

    \centering

  \renewcommand{\arraystretch}{1.4}
    \resizebox{\textwidth}{!}{

  \begin{tabular}{c|cccccccccccc|c}
\hline
        \textbf{Method}
        & \textbf{Ar}$\rightarrow$\textbf{Cl} & \textbf{Ar}$\rightarrow$\textbf{Pr} & \textbf{Ar}$\rightarrow$\textbf{Rw} & \textbf{Cl}$\rightarrow$\textbf{Ar} & \textbf{Cl}$\rightarrow$\textbf{Pr} & \textbf{Cl}$\rightarrow$\textbf{Rw} & \textbf{Pr}$\rightarrow$\textbf{Ar} & \textbf{Pr}$\rightarrow$\textbf{Cl} & \textbf{Pr}$\rightarrow$\textbf{Rw} & \textbf{Rw}$\rightarrow$\textbf{Ar} & \textbf{Rw}$\rightarrow$\textbf{Cl} & \textbf{Rw}$\rightarrow$\textbf{Pr} & \textbf{Avg} \\
        \hline
No Adaptation & 44.6 & 62.7 & 72.0 & 52.1 & 62.7 & 65.1 & 52.9 & 43.0 & 73.9 & 63.7 & 45.8 & 77.3 & 59.7 \\  
 DeepJdot~\cite{bhushan2018deepjdot} & 39.7 & 50.4 & 62.5 & 39.5 & 54.4 & 53.2 & 36.7 & 39.2 & 63.5 & 52.3 & 45.4 & 70.5 & 50.6 \\
DAN~\cite{DAN}  & 43.6 & 57.0  & 67.9 & 45.8 & 56.5 & 60.4 & 44.0 & 43.6 & 67.7 & 63.1 & 51.5 & 74.3 & 56.3\\
DANN~\cite{cite:JMLR16RevGrad} & 45.6 & 59.3  &70.1 &47.0 & 58.5  &60.9 & 46.1& 43.7& 68.5& 63.2& 51.8& 76.8& 57.6\\
JAN~\cite{JAN}  &45.9 &61.2 &68.9 &50.4 &59.7 &61.0 &45.8 &43.4 &70.3 &63.9 &52.4 &76.8 &58.3\\
CDAN~\cite{CDANLong} & \textbf{50.7} & \textbf{70.6} & \textbf{76} & \mbox{\bf \em 57.6} & \textbf{70} & \mbox{\bf \em 70} & \mbox{\bf \em 57.4} & \textbf{50.9} & \textbf{77.3} & \textbf{70.9} & \textbf{56.7} & \textbf{81.6} & \textbf{65.8} \\   
\hline
Ours  &\mbox{\bf \em  49.9$\pm$0.2} & \mbox{\bf \em 68.6$\pm$0.3} & \mbox{\bf \em 74.68$\pm$0.5} & \textbf{59.9$\pm$0.4} & \mbox{\bf \em 68.92$\pm$0.6} & \textbf{71.82$\pm$0.2} & \textbf{58.12$\pm$0.1} & \mbox{\bf \em 49.4$\pm$0.1} & \textbf{77.3$\pm$0.2} & \mbox{\bf \em 68.7$\pm$0.3} & \mbox{\bf \em 54.82$\pm$0.2} & \mbox{\bf \em 78.92$\pm$0.3} & \mbox{\bf \em 65.1$\pm$0.4} \\ \hline
\end{tabular}}

\caption{\textbf{UDA performance on the Office-Home dataset}. Best performance is shown in {\bf bold}, and the second best in {\bf \em bold italic}.}
\label{table:officehome}
\end{table*}

\noindent\textbf{Model}:
In accordance with the prior literature, we choose the pre-trained ResNet-50~\cite{ResNet} as our feature extractor and as in previous case, we fine tune it to obtain the $2048$-dimensional features and set the subspace dimension as $800$. 
The classifier and subspace alignment networks are trained with the same hyper-parameters as in Section \ref{sec-imageclef}. From Table \ref{table:visda}, it is interesting that subspace alignment leads to more than 3\% improvements over advanced adversarial learning methods such as CDAN~\cite{CDANLong}. Figure \ref{fig:tsne} visualizes the t-SNE embeddings of the features before and after performing UDA. As expected, the class boundaries become very well separated through our proposed subspace alignment strategy. 



\subsection{Office-Home}

\noindent\textbf{Datasets}: This challenging dataset~\cite{cite:CVPR17DHN} contains 15,500 images in 65 classes from office and home settings, forming $4$ extremely dissimilar domains: Artistic images (\textbf{Ar}), Clip Art (\textbf{Cl}), Product images (\textbf{Pr}), and Real-World images (\textbf{Rw}).

\noindent\textbf{Model}: Similar to Section \ref{sec-imageclef}, we train TAFE with a pre-trained ResNet-50 as our backbone and obtain the $2048$-dimensional features, and build subspaces of $800$ dimensions. The classifier and auxiliary networks are trained with the same hyper-parameters as earlier. 
Comparisons to the state-of-the-art methods can be found in Table \ref{table:officehome}. We observe that while our approach consistently outperforms baseline methods including methods such as DeepJdot~\cite{bhushan2018deepjdot}, with comparable performance to the highest reported -- CDAN \cite{CDANLong} in terms of the average accuracy across all pairs of DA tasks.


 \subsection{Data and Parameter Efficiency}
 \label{sec:less data }
\begin{figure}[t]
    \centering
    \includegraphics[width = 0.8\columnwidth]{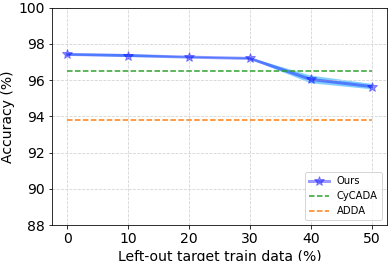}
    \caption{\textbf{Impact of target data size}. Performance on the target test-set with varying sizes of target training data on the USPS to MNIST adaptation task. The baseline results with ADDA and CyCADA were obtained using the entire training set.}
    \label{fig:reduced_data}
\end{figure}

  \noindent\textbf{Data Efficiency}: Owing to its design simplicity, we surmise that our approach admits improved data efficiency. 
 To test this hypothesis, we evaluate our method under the scenario where the amount of unlabeled target data is limited. While we train TAFE, classifier and subspace alignment networks with varying target data sizes, we report accuracies on the full target test set. In particular, we consider the problem of adapting USPS $\rightarrow$ MNIST and the results from $3$ random trials are shown in Figure \ref{fig:reduced_data}. It can be seen that even with $30\%$ lesser training data in the target domain, our approach still outperforms state-of-the-art baselines that have access to the entire data. Further, the drop in performance even when operating at only 50\% of data is very low ($\approx$ 2\% points), thus evidencing that a simpler alignment strategy can reduce the data requirements while not compromising the performance.\\
 \noindent\textbf{Parameter Efficiency}:
 As an another important consequence of design simplicity,
our approach leads to large parameter savings. As an example for the VISDA dataset, CDAN on top of the base ResNet50 feature extractor ($23$M parameters) requires an additional discriminator network ($>4$M parameters) while our approach on top of ResNet50 requires only one linear layer for adaptive subspace alignment ($<700$K parameters) thus registering a savings of $4$M parameters. In fact, our approach almost always requires much less parameters compared to adversarial domain adaptation methods such as CyCADA~\cite{CYCADA}, CDAN~\cite{CDANLong}, GTA~\cite{GTA}, UNIT~\cite{UNIT} etc.
Since our approach entirely dispenses the need for adversarial learning, we also remove its inherent training instabilities.  
Also note that, once TAFE is trained, in the alignment stage we do not update any parameters except for classifier and subspace alignment layers, thus reducing the explicit alignment time considerably. Furthermore, in the next section, we show that our simple design and procedure is robust and leads to better performance even in the very challenging partial domain adaptation setting when compared to existing UDA methods.

 \subsection{Role of Subspace Dimensionality}
\begin{figure}[t]
    \centering
    \includegraphics[width=0.8\columnwidth]{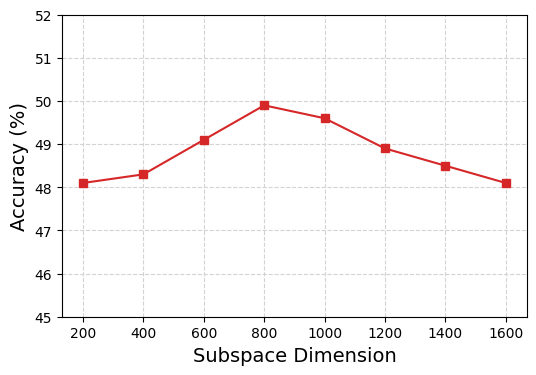}
    \caption{\textbf{Role of subspace dimensionality}. UDA performance for the case Art $\rightarrow$ Clipart from the Office-Home dataset as a function of subspace dimensionality.}
    \label{fig:susbspace_dim}
\end{figure}

In Figure~\ref{fig:susbspace_dim}, we plot the mean adaptation accuracy over $5$ trials on the OfficeHome benchmark with source domain Art and target domain Clipart. It can be seen that the accuracy is relatively stable with respect to choice of subspace dimensions and with the best performance at $d=800$. Note, we set this hyper-parameter at $800$ for all our experiments.  

\subsection{Results on Partial Domain Adaptation}
\label{sec:PDA}
 \begin{table*}[t]

    \centering
    \renewcommand{\arraystretch}{1.4}
    \resizebox{\textwidth}{!}{
    \begin{tabular}{c|cccccccccccc|c}
    	\hline
        \textbf{Method}
        & \textbf{Ar}$\rightarrow$\textbf{Cl} & \textbf{Ar}$\rightarrow$\textbf{Pr} & \textbf{Ar}$\rightarrow$\textbf{Rw} & \textbf{Cl}$\rightarrow$\textbf{Ar} & \textbf{Cl}$\rightarrow$\textbf{Pr} & \textbf{Cl}$\rightarrow$\textbf{Rw} & \textbf{Pr}$\rightarrow$\textbf{Ar} & \textbf{Pr}$\rightarrow$\textbf{Cl} & \textbf{Pr}$\rightarrow$\textbf{Rw} & \textbf{Rw}$\rightarrow$\textbf{Ar} & \textbf{Rw}$\rightarrow$\textbf{Cl} & \textbf{Rw}$\rightarrow$\textbf{Pr} & \textbf{Avg} \\
        \hline
        ResNet~\cite{cite:NIPS12CNN} & 38.57 & 60.78 & 75.21 & 39.94 & 48.12 & 52.90 & 49.68 & 30.91 & 70.79 & 65.38 & 41.79 & 70.42 & 53.71 \\
        DAN~\cite{DAN} & 44.36 & 61.79 & 74.49 & 41.78 & 45.21 & 54.11 & 46.92 & 38.14 & 68.42 & 64.37 & 45.37 & 68.85 & 54.48 \\
        DANN~\cite{cite:ICML15RevGrad} & 44.89 & 54.06 & 68.97 & 36.27 & 34.34 & 45.22 & 44.08 & 38.03 & 68.69 & 52.98 & 34.68 & 46.50 & 47.39 \\
        RTN~\cite{cite:NIPS16RTN} & {{49.37}} & {{64.33}} & 76.19& {{47.56}} &{{ 51.74}} & {{57.67}} & {{50.38}} & {{41.45}} & \textbf{\textit{75.53}} & {{70.17}} & {{51.82}} & {{74.78}} & {{59.25}} \\

    	PADA & \textbf{51.95} & \textbf{67} & \textbf{78.74} & \textit{\textbf{52.16}} & \textbf{\textit{53.78}} & \textbf{\textit{59.03}} & \textbf{\textit{52.61}} & \textbf{\textit{43.22}} & \textbf{78.79} & \textbf{73.73} & \textbf{56.6} & \textbf{\textit{77.09}} & \textbf{\textit{62.06}} \\
    	
    	\hline
    	
    	{Ours} & \textit{\textbf{51.1}} & \textit{\textbf{65.93}} & \textit{\textbf{76.42}} & \textbf{65.01} & \textbf{65.76} & \textbf{72.77} & \textbf{64.18} & \textbf{49.31} & 74.60 & \textit{\textbf{70.43}} & \textit{\textbf{54.9}} & \textbf{77.3} & \textbf{65.64}\\
    	
    	\hline

    \end{tabular}
      
    }

    \caption{\textbf{Performance on partial domain adaptation}. Classification accuracy on partial domain adaptation tasks from \emph{Office-Home} (ResNet-50). Note only PADA is a PDA baseline, while all other baselines general-purpose UDA methods.}
    \label{tab:pda_officehome}
\end{table*}

Partial Domain Adaptation (PDA)~\cite{cao2018partial} is a recently proposed, challenging domain adaptation scenario. In this setting, the categories in the target domain can be a subset of the categories observed in the source domain, i.e., $\mathcal{Y}_{t} \subset \mathcal{Y}_{s}$. Many standard closed-set domain adaptation methods with explicit alignment techniques suffer a drastic drop in performance when tested in this setting.
This is because, samples from the missing classes lead to negative transfer of knowledge. We investigate the robustness of the proposed approach in this extreme setting.

We conduct experiments on the popular OfficeHome PDA benchmark proposed in~\cite{cao2018partial}. For this benchmark, we use images from the first 25 classes in alphabetical order as the target domain and images from all 65 classes as the source domain. We make use of the same splits and experimental protocol provided\footnote{https://github.com/thuml/PADA} by~\cite{cao2018partial} for a fair comparison. 

We compare against standard domain adaptation methods along with PADA~\cite{cao2018partial}, which has been specifically designed to work in this setting.
In order to adapt our proposed approach to PDA, we make only one change to our algorithm - modify $\mathcal{L}_{c}$ to account for the reduced number of classes in the target. We report the results of our approach in Table~\ref{tab:pda_officehome}. From the results, it can be seen that, even under this highly challenging scenario,
the proposed approach performs on par with PADA and significantly outperforms other UDA baselines. Using subspace-based alignment produces inherently well-regularized models, thus improving their robustness to both domain and label shifts.



\subsection{Results on Progressive Domain Adaptation}
\begin{table*}[t]

    \centering

  \renewcommand{\arraystretch}{1.4}
    \resizebox{\textwidth}{!}{

  \begin{tabular}{c|cccccccccccc|c}
\hline
        \textbf{Method}
        & \textbf{Ar}$\rightarrow$\textbf{Cl} & \textbf{Ar}$\rightarrow$\textbf{Pr} & \textbf{Ar}$\rightarrow$\textbf{Rw} & \textbf{Cl}$\rightarrow$\textbf{Ar} & \textbf{Cl}$\rightarrow$\textbf{Pr} & \textbf{Cl}$\rightarrow$\textbf{Rw} & \textbf{Pr}$\rightarrow$\textbf{Ar} & \textbf{Pr}$\rightarrow$\textbf{Cl} & \textbf{Pr}$\rightarrow$\textbf{Rw} & \textbf{Rw}$\rightarrow$\textbf{Ar} & \textbf{Rw}$\rightarrow$\textbf{Cl} & \textbf{Rw}$\rightarrow$\textbf{Pr} & \textbf{Avg} \\
        \hline
No adaptation& 44.4 & 61.95 & 65.8 & 55.5 & 65.7 & 67.05 & 52.35 & 42.65 & 69.85 & 57.9 & 47.9 & 73.5 & 58.71 \\  
Progressive & \mbox{\bf \em 48.42} & \textbf{69.72} & \mbox{\bf \em 73.4} & \mbox{\bf \em 59.1 }& \textbf{77.0} & \textbf{73.5} & \mbox{\bf \em 56.1} & \mbox{\bf \em 48.4} & \mbox{\bf \em 77.1} &\mbox{\bf \em  60.7} & \mbox{\bf \em 50.85} & \textbf{79.69} & \textit{\textbf{64.5}} \\
Retraining  & \textbf{49.9} & \mbox{\bf \em 68.6 } & \textbf{74.7} & \textbf{59.9} & 68.92 & 71.82 & \textbf{58.12} & \textbf{49.4} & \textbf{77.3} & \textbf{68.7} & \textbf{54.82} & \mbox{\bf \em 78.92} & \textbf{65.1}\\
\hline
\end{tabular}}
\caption{\textbf{Progressive adaptation results}. Here, we report the results on generalizing to a new unlabeled target domain at test-time using the proposed progressive adaptation. Best performance is shown in {\bf bold}, and the second best in {\bf \em bold italic}.}
\label{tab:progressiveDA}

\end{table*}

As described in Section~\ref{sec:progressive DA}, we extend our approach to another challenging setting, which requires an already deployed DA model to adapt to a \textit{new unlabeled target domain} at test-time. In Table~\ref{tab:progressiveDA}, we show the results under this setting for the Office-Home dataset. The results reported for each case, $\mathcal{D}_s \rightarrow \mathcal{D}_A \rightarrow \mathcal{D}_B$, were obtained as an average of accuracies from two different choices for $D_A$. For example, the performance for \textit{Ar $\rightarrow$ Cl} is the average performances obtained using two independent progressive adaptation runs, namely \textit{Ar $\rightarrow$ Pr $\rightarrow$ Cl} and \textit{Ar $\rightarrow$ Rw $\rightarrow$ Cl}. For comparison, we show the results obtained without any test-time adaptation to the new target domain and by retraining the models (TAFE and the alignment steps) from scratch (results from Table \ref{table:officehome}).

As can be seen from the results, without needing to even fine-tune the feature extractor and by only recomputing the parameters of the classifier and the alignment matrix, we observe an average improvement of $6$\% over directly using the deployed model without any test-time adaptation. Furthermore, the performance is on par with models that are retrained from scratch for every scenario. This result clearly evidences the flexible nature of the proposed approach and validates our key hypothesis that, with powerful feature extractors and a carefully tailored optimization strategy, subspace alignment is highly effective for UDA.

\section{Conclusions}
In this work, we presented a principled and effective approach to tackle the problem of unsupervised domain adaptation, in the context of visual recognition. 

Our work revisits the traditionally popular solution of subspace-based DA, in the wake of recent successes in deep representation learning. The proposed method relies on pre-training a target-aware feature extractor and poses domain alignment as an auxiliary task to the primary task of learning a well-calibrated classifier for both source and target domains. Through an alternating optimization strategy, wherein the pre-trained feature extractor is frozen, our approach effectively balances the domain alignment and model fidelity objectives. Using rigorous empirical studies, we showed that our method achieves competitive or sometimes higher performance than the state-of-the-art methods on several UDA benchmarks. Since our approach produces intrinsically well-regularized models, we demonstrated that it provides strong boosts in performance on the challenging partial domain adaptation setting. Furthermore, as our approach isolates feature learning and alignment steps, we found that our approach can adapt to new domains progressively at test-time, without requiring to train the feature extractor from scratch. Owing to the simplicity of our approach, we also demonstrated a significant reduction in the target data and computational requirements. 
Future extensions to this work includes incorporating self-training protocols, and extending our methodology to tasks such as semantic segmentation, open-set classification~\cite{Saito_2018_ECCV}, and image-to-image translation.

\label{Conclusion}

\section*{Acknowledgements}
This work was performed under the auspices of the U.S. Department of Energy by the Lawrence Livermore National Laboratory under Contract No. DE-AC52-07NA27344, Lawrence Livermore National Security, LLC.

\appendices

\ifCLASSOPTIONcaptionsoff
  \newpage
\fi

\bibliographystyle{unsrt}
\bibliography{references}
\end{document}